\documentclass[letterpaper, 10 pt, conference]{ieeeconf}  

\IEEEoverridecommandlockouts                              

\usepackage{graphics} 
\usepackage{epsfig} 
\usepackage{mathptmx} 
\usepackage{times} 
\usepackage{amsmath} 
\usepackage{amssymb}  
\usepackage{subfigure}
\usepackage{color}
\usepackage{colortbl}
\usepackage{multirow}
\usepackage{makecell}
\usepackage{cite}
\usepackage{hyperref}
\hypersetup{colorlinks}
\let\labelindent\relax
\usepackage{enumitem}
\usepackage{dsfont}
\usepackage{xurl}
\usepackage{siunitx}
\usepackage{cleveref}
\usepackage{capt-of}

\usepackage{url}
\definecolor{cite}{RGB}{65,105,225}

\crefname{figure}{Fig.}{Figs.}
\crefname{section}{Sec.}{Secs.}
\crefname{equation}{Eq.}{Eqs.}

\usepackage{booktabs}
\newcommand{\ci}[1]{\tiny{\textcolor{gray}{~($\pm #1$})}}

\PassOptionsToPackage{dvipsnames,table}{xcolor}
\usepackage{xcolor}
\usepackage{graphicx}
\usepackage{booktabs}
\usepackage{amsmath}
\usepackage{multirow}
\usepackage{makecell}
\usepackage{csquotes}
\usepackage{wrapfig}
\usepackage{diagbox}

\usepackage[utf8]{inputenc}
\usepackage{color,soul}
\usepackage[T1]{fontenc}

\definecolor{trashbin_blue}{RGB}{23,211,253}
\definecolor{hydrant_green}{RGB}{37,253,53}
\definecolor{bench_orange}{RGB}{255,195,45}

\setul{0.5ex}{0.3ex}
\setulcolor{trashbin_blue}

\usepackage{CJKutf8}

\definecolor{mygray}{RGB}{230,230,230}

\definecolor{clipcolor}{gray}{1.0}

\definecolor{openclipcolor}{gray}{1.0}

\definecolor{evaclipcolor}{gray}{1.0}

\definecolor{siglipcolor}{gray}{1.0}

\definecolor{llavacolor}{gray}{1.0}

\definecolor{sizescolor}{RGB}{194,215,255}

\definecolor{sizemcolor}{RGB}{165,202,255}

\definecolor{resscolor}{RGB}{194,255,191}

\definecolor{resmcolor}{RGB}{152,243,165}

\usepackage[most]{tcolorbox}

\colorlet{mapcolor}{ForestGreen}
\colorlet{langcolor}{RoyalBlue}
\colorlet{visioncolor}{WildStrawberry}
\colorlet{colabcolor}{YellowOrange}

\newtcbox{\cvtag}{enhanced,nobeforeafter,tcbox raise base,boxrule=0.4pt,top=0mm,bottom=0mm,
  right=0mm,left=4mm,arc=2pt,boxsep=2pt,before upper={\vphantom{dlg}},
colframe=visioncolor!50!black,coltext=visioncolor!25!black,colback=visioncolor!10!white,
  overlay={\begin{tcbclipinterior}\fill[visioncolor!80] (frame.south west)
    rectangle node[text=white,font=\sffamily\bfseries\tiny,rotate=90] {CV} ([xshift=4mm]frame.north west);\end{tcbclipinterior}}}

\newtcbox{\sharetag}{enhanced,nobeforeafter,tcbox raise base,boxrule=0.4pt,top=0mm,bottom=0mm,
  right=0mm,left=4mm,arc=2pt,boxsep=2pt,before upper={\vphantom{dlg}},
colframe=langcolor!50!black,coltext=langcolor!25!black,colback=langcolor!10!white,
  overlay={\begin{tcbclipinterior}\fill[langcolor!80] (frame.south west)
    rectangle node[text=white,font=\sffamily\bfseries\tiny,rotate=90] {TRA} ([xshift=4mm]frame.north west);\end{tcbclipinterior}}}

\newtcbox{\navitag}{enhanced,nobeforeafter,tcbox raise base,boxrule=0.4pt,top=0mm,bottom=0mm,
  right=0mm,left=4mm,arc=2pt,boxsep=2pt,before upper={\vphantom{dlg}},
colframe=colabcolor!50!black,coltext=colabcolor!25!black,colback=colabcolor!10!white,
  overlay={\begin{tcbclipinterior}\fill[colabcolor!80] (frame.south west)
    rectangle node[text=white,font=\sffamily\bfseries\tiny,rotate=90] {NAV} ([xshift=4mm]frame.north west);\end{tcbclipinterior}}}

\newtcbox{\loctag}{enhanced,nobeforeafter,tcbox raise base,boxrule=0.4pt,top=0mm,bottom=0mm,
  right=0mm,left=4mm,arc=2pt,boxsep=2pt,before upper={\vphantom{dlg}},
colframe=mapcolor!50!black,coltext=mapcolor!25!black,colback=mapcolor!10!white,
    overlay={\begin{tcbclipinterior}\fill[mapcolor!80] (frame.south west)
        rectangle node[text=white,font=\sffamily\bfseries\tiny,rotate=90] {LOC} ([xshift=4mm]frame.north west);\end{tcbclipinterior}}}

\newcommand{\shareul}[1]{\setulcolor{langcolor}\ul{#1}}
\newcommand{\navul}[1]{\setulcolor{colabcolor}\ul{#1}}
\newcommand{\locul}[1]{\setulcolor{mapcolor}\ul{#1}}

\tcbuselibrary{skins}

\AtBeginEnvironment{tcolorbox}{\footnotesize}

\usepackage[noorphans,vskip=1em,leftmargin=1em]{quoting}

\title{\LARGE \bf
Unified Humanoid Fall-Safety Policy from a Few Demonstrations

}

\author{Zhengjie Xu, Ye Li, Kwan-Yee Lin, Stella X. Yu \\[0.2cm]
University of Michigan\\
{\tt\small \{zhengjie, yeyli, junyilin, stellayu\}@umich.edu}}

\begin{document}

\twocolumn[{%
\renewcommand\twocolumn[1][]{#1}%
\maketitle

\begin{center}
    \centering
    \vspace{-0.5cm}
    \includegraphics[width=\linewidth]{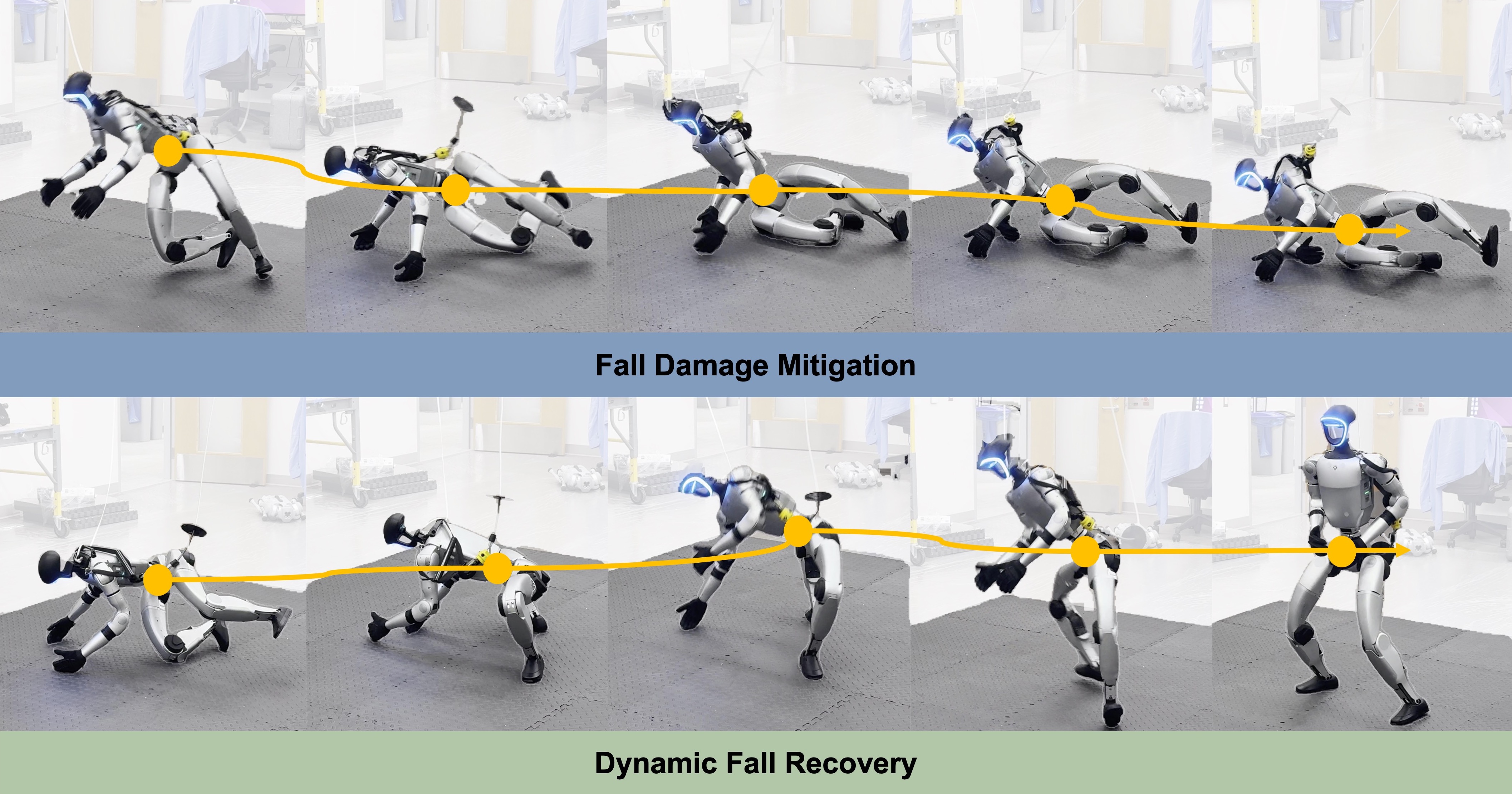}
    \captionof{figure}{{\bf Our method enables humanoids to fall safely and rise promptly.} 
Snapshots show real-world deployment on the Unitree G1: When suddenly destabilized, the robot redirects into a side fall with arm buffering, then reorients and rises, demonstrating adaptive and resilient recovery.}
    \label{fig:teaser-ph}
\end{center}
}]

\thispagestyle{empty}
\pagestyle{plain}


\begin{abstract}
Falling is an inherent risk of humanoid mobility. Maintaining stability is thus a primary safety focus in robot control and learning, yet no existing approach fully averts loss of balance.
When instability does occur, prior work addresses only isolated aspects of falling: avoiding falls, choreographing a controlled descent, or standing up afterward.  Consequently, humanoid robots lack integrated strategies for impact mitigation and prompt recovery when real falls defy these scripts.
We aim to go beyond keeping balance to make the entire fall-and-recovery process safe and autonomous: prevent falls when possible, reduce impact when unavoidable, and stand up when fallen. By fusing sparse human demonstrations with reinforcement learning and an adaptive diffusion-based memory of safe reactions, we learn adaptive whole-body behaviors that unify fall prevention, impact mitigation, and rapid recovery in one policy. Experiments in simulation and on a Unitree G1 demonstrate robust sim-to-real transfer, lower impact forces, and consistently fast recovery across diverse disturbances, pointing toward safer, more resilient humanoids in real environments.  Videos are available at~\url{https://firm2025.github.io/}.

\end{abstract}

\section{Introduction}

Where there are legs, there will be stumbles. Even the most carefully trained humanoids - built for agile locomotion and intelligent navigation planning - are bound to be jolted off balance by a stray push, a loose stone, or an unexpected gust. When a 1.3\,m, 35\,kg Unitree G1 robot with delicate vision and force sensors topples, the damage can be costly: bent joints, cracked housings, and extended downtime.

Such incidents are not rare anomalies but fundamental risks of legged mobility. 
Balance controllers can reduce but never eliminate unexpected falls~\cite{kajita2003zmp,stephens2007push}. 
Unlike wheeled or quadruped robots, which enjoy wider and more stable support base~\cite{di2018dynamic,ma2023learning}, 
humanoids combine tall, narrow frames with dozens of degrees of freedom, producing diverse and hard-to-predict fall dynamics~\cite{fujiwara2003first,rossini2019optimal}.

{\bf We aim to give humanoids a single instinct for self-preservation: a unified policy that keeps them upright whenever possible and, when a fall is unavoidable, ensures they fall safely and rise on their own} (Fig.~\ref{fig:teaser-ph}).

Prior work tackles only isolated pieces of this chain.  
Classical balance controllers focus on avoiding falls altogether~\cite{kajita2003zmp,stephens2007push},  
motion-planning methods choreograph a controlled descent~\cite{rossini2019optimal,wang2017real},  
and recovery studies begin only after the damage is done, teaching robots to stand up from static supine postures~\cite{huang2025learning,he2025learning}.

{\bf Yet falling and rising are inseparable phases of a single physical process}: How a robot falls directly shapes how it can get back up. By unifying mitigation and recovery, our approach explicitly addresses this coupled dynamic.

The challenge is daunting. Once balance is lost, a fall becomes a complex, high-dimensional physical process with rapidly changing contacts and forces, exposing weaknesses in both major camps of humanoid control:
\begin{enumerate}[leftmargin=*,itemsep=0pt,topsep=0pt]
\item{\bf Model-based control.}  
Carefully planned motions can be computed for particular impacts~\cite{wang2017real,rossini2019optimal},  
but such methods depend on simplified dynamics and become intractable as the range of disturbances grows.
\item{\bf Learning-based control.}  
Imitation learning typically requires dense, full-motion demonstrations, which are difficult to collect at scale and often lead to policies that collapse to fixed reference trajectories with poor adaptability~\cite{DBLP:journals/corr/abs-2502-01465}.  
Reinforcement learning (RL) must juggle a set of carefully crafted reward terms whose interactions are hard to anticipate, making reward engineering difficult and often producing brittle or unnatural behaviors~\cite{hwangbo2019learning,chen2024learning}.  
Without an effective way to represent multi-modal policies (e.g., through skill embeddings or generative models), RL struggles to encode the diverse actions needed for safe falling and rising~\cite{peng2022ase,chi2023diffusion}.
\end{enumerate}
Due to these limitations, no prior method reliably spans the full spectrum from balance maintenance, through damage-mitigating fall, to autonomous recovery.

We tackle this challenge with learning a single, unified humanoid fall-safety policy from just a few demonstrations.  By fusing sparse human demonstrations with reinforcement learning and an adaptive diffusion-based memory of safe reactions, we learn adaptive whole-body behaviors that cover fall prevention, impact mitigation, and rapid recovery within a single policy (Fig.~\ref{fig:figure2}). The policy learning proceeds in two stages: \textit{learning safe skill priors} and \textit{learning adaptiveness}, achieved through the following four steps:
\begin{enumerate}[leftmargin=*,itemsep=0pt,topsep=0pt]
\item \textbf{Seed safe skill acquisition.}  
The robot begins with a few temporally sparse human key poses, internalizing them through RL to fit its own morphology and dynamics.  This creates dense reaction trajectories that seed safe falling and rising in its action space.

\item \textbf{Safe skill enrichment.}  
Targeted stitching of compatible falling and rising motions, combined with policy roll-outs, generates additional safe trajectories. This expansion yields strategies for pre-emptive fall prevention, diverse fall variations, fall mitigation, and reliable recovery.

\item \textbf{Safe reactive memory.}  
All safe reactions are distilled into a diffusion policy that captures a rich, multi-modal distribution of fall-and-rise behaviors.  A learned feature predicts the next safe target pose from past trajectory data.

\item \textbf{Adaptive safe control.}  
At run time, the feature is extracted online to retrieve the nearest neighbour from a memory bank of safe poses.  
Refreshing predictions at every step, the system assembles safe trajectories on the fly from overlapping segments, expanding each target into a neighborhood of possibilities and enabling rapid adaptation to unforeseen terrain or disturbances.
\end{enumerate}
The result is a humanoid that does more than stay on its feet.  It anticipates trouble, redirects unavoidable falls to minimize harm, and rebounds swiftly to a stable stance, turning an inevitable weakness into evidence of genuine resilience.  

Experiments in simulation and on the Unitree G1 confirm robust sim-to-real transfer, with lower impact forces and prompt, reliable recovery across diverse disturbances~\cite{hwangbo2019learning,he2025learning,huang2025learning}.  
By unifying pre-emptive fall prevention, impact mitigation, and rapid recovery within a single memory-driven policy, our approach advances safe humanoid control and lays a strong foundation for resilient service and assistive robots in unstructured environments.

\section{Related Work}

\subsection{Humanoid Control}

Model-based methods laid the foundations of humanoid control~\cite{di2018dynamic,chignoli2021humanoid}.  
Learning-based methods have since advanced the field, from IL~\cite{liu2024opt2skill} to RL~\cite{hwangbo2019learning, chen2024learning}.  
Human demonstrations further enrich motion style and diversity~\cite{peng2018deepmimic,peng2022ase,he2024omnih2o,haarnoja2024learning}.  
Recent studies extend locomotion to challenging motions and diverse terrains~\cite{zhuang2024humanoid,DBLP:conf/cvpr/LinY25,cheng2024expressive,gu2024advancing,li2025reinforcement,radosavovic2024learning}.  
Our work complements these advances with a unified learning framework that goes beyond locomotion to prevent falls, mitigate impact, and recover robustly without heavy reward engineering.

\subsection{Humanoid Fall Mitigation}

Early methods imitate human break-falls to limit damage~\cite{fujiwara2002falling,fujiwara2003first,fujiwara2004safe,meng2017falling}, but rely on heuristics and offline tuning. 
Model-based methods cast safe falling as momentum redirection, trajectory optimization, or multi-contact planning~\cite{lee2011fall,wang2012whole,wang2017real,wang2018unified}, and energy-based controllers provide online shaping~\cite{subburaman2018energy,subburaman2018online}. 
Mechanical or control compliance lengthens impact and regulates post-impact behavior~\cite{zhang2017biomimetic,luo2016biped}, while direction-control strategies steer the body toward safer contact regions~\cite{yun2009safe,goswami2014direction}. 
These methods work in targeted scenarios but typically depend on hand-crafted strategies, simplified dynamics, or pre-specified contact sequences; in contrast, our work learns a unified policy that generalizes across disturbances, enabling pre-emptive fall prevention, impact mitigation, and prompt recovery within one framework.

\subsection{Humanoid Fall Recovery}

Classical model-based approaches plan stand-up motions after a fall~\cite{fujiwara2003first, fujiwara2004safe, stuckler2006getting, kanehiro2007getting}, but they generalize poorly and are sensitive to disturbances. 
Learning-based methods improve robustness: Some imitate predefined trajectories~\cite{peng2018deepmimic,yang2023learning,peng2022ase}, while others train policies from scratch~\cite{tao2022learning,huang2025learning,he2025learning}. 
Although these works broaden the range of recoverable postures, resulting motions often remain unnatural and fragile. 
A recent quadruped study jointly addressed falling and recovery~\cite{ma2023learning}, but no prior humanoid work unifies fall mitigation and prompt recovery. Our work closes this gap by integrating all into a single policy.

\subsection{Diffusion Models in Robotics}

Diffusion models have recently been adopted for control and planning by casting policy learning as conditional generative modeling~\cite{ajayconditional,janner2022planning,chi2023diffusion}. 
Building on these foundations, legged-robot studies learn multi-skill policies from offline data and deploy them online. 
For example, DiffuseLoco achieves robust zero-shot transfer for quadruped locomotion~\cite{huang2025diffuseloco}, while preference alignment and test-time guidance improve robustness in out-of-distribution states~\cite{yuan2024preference}. 
Hybrid approaches embed MPC for constraint satisfaction and safety~\cite{zhoudiffusion,romer2025diffusion}. 
Our work leverages diffusion to encode a multi-modal memory of safe fall-and-rise behaviors.

\begin{figure*}[t]\centering
\includegraphics[width=\linewidth]{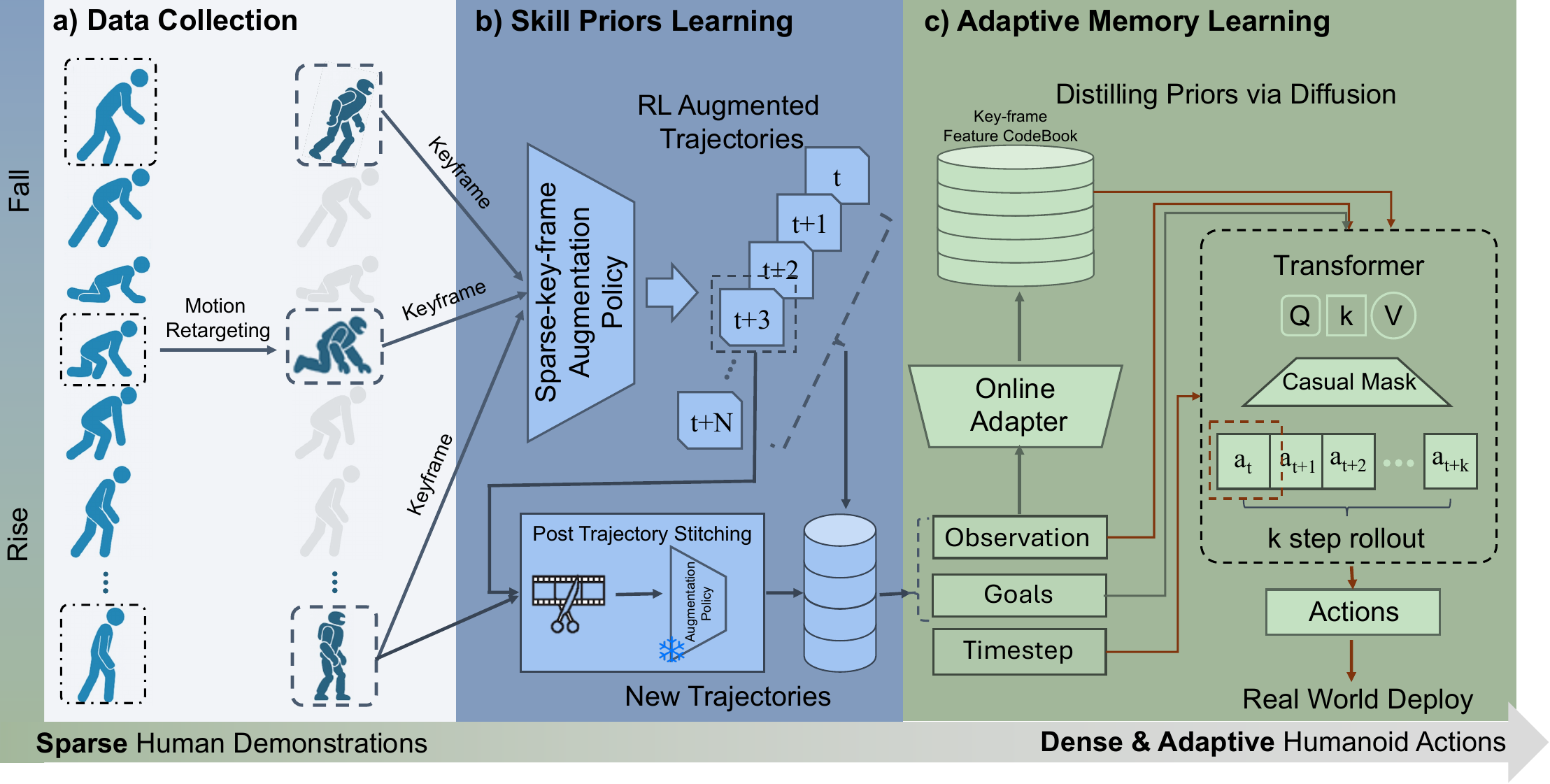}\\[-6pt]
\caption{{\bf Workflow Overview}. From a few sparse human key poses, the robot seeds safe fall–rise skills, expands them through RL with post-trajectory stitching, distills enriched behaviors into a diffusion-based action memory, and composes online adapter to execute actions with context-awareness.
}\vspace{-10pt}
\label{fig:figure2}
\end{figure*}
\section{Problem Formulation}

We study the problem of \textit{fall damage mitigation and recovery} for humanoid robots in unstructured environments, formulated as a \textit{dynamic process} that begins with a destabilizing disturbance to fall and ends once the robot regains a stable upright pose above a target height. This process inherently involves two levels: 1) \textit{damage mitigation} (minimizing impact during the fall), and 2) \textit{recovery} (standing back up)---but unlike prior works, we do not separate them; instead, we learn the coupled process directly. At each timestep $t$, we perceive the robot proprioception information to feed into the policy to output action $a_t\in \mathbb{R}^{23}$, which are offset applied to the robot’s nominal joint configuration $q^{\text{default}}$. By learning this unified dynamic process, a single policy can be directly applied to three tasks: \textit{fall mitigation only}, \textit{recovery only}, and the \textit{full coupled problem}, without task-specific retraining.

\section{Method}

We present \textit{FIRM}, (short for \underline{f}all m\underline{i}tigation and \underline{r}ecovery from a few human de\underline{m}onstrations), a control policy for the diverse, complex dynamics of humanoid falling and recovery (Fig.~\ref{fig:figure2}).  
FIRM unifies fall mitigation and recovery in a single framework that balances \textit{safety} and \textit{behavioral adaptiveness}.  It operates in two stages.  
\textbf{1) Skill priors} (Sec.~\ref{method-prior}): a few human demonstrations are fitted and retargeted to the G1 humanoid, then sparsified into key frames (Fig.~\ref{fig:figure2}a).  
These seed skills are expanded with RL-based augmentation and post-stitching (Fig.~\ref{fig:figure2}b) to produce diverse, damage-reducing trajectories.  
\textbf{2) Adaptive memory} (Sec.~\ref{method-adapt}): the enriched trajectories are distilled into a diffusion model and paired with a lightweight adapter (Fig.~\ref{fig:figure2}c) that enables real-time fall mitigation and recovery across varied conditions.

\subsection{Learning Fall-and-Recover Skill Priors}\label{method-prior}

\subsubsection{Collecting Seed safe skill via Retargeting Human Videos}

We expect the task controller to prioritize safety during falling and recovery, which requires safe motion patterns. Since such patterns are difficult to engineer manually, we utilize human demonstrations, which naturally encode safety-critical behaviors. However, current large-scale MoCap datasets like AMASS~\cite{AMASS:ICCV:2019} lack realistic fall–recovery motions. To address this gap, we collect a small number of monocular human video demonstrations, and process them through fitting to SMPL~\cite{DBLP:journals/tog/LoperM0PB15} and retargeting to the G1 humanoid. In total, we use $5 (2:2:1)$ high-quality trajectories covering forward, sideways, and backward fall-recover processes on flat ground, collected from subjects of different genders and heights to provide variety in motion styles. 

While the data volume of demonstrations is small in scale, its quality and targeted varieties are crucial, as they provide sufficient prototypes for our later designs to compose and expand upon. Each trajectory includes the information of all joint positions $q_i$, rigid body poses relative to root $T_{b_i}$ and root poses $T_{\text{root}}$ retargetted to G1 robot. To obtain velocities, we further calculate joint velocities $\dot{q}_i$, rigid body twists relative to root $V_{b_i}$, and root twist $V_{\text{root}}$ using finite difference.

\subsubsection{Expanding Priors via Sparse-key-frame Augmentation Policy Learning}
Unlike conventional motion-tracking tasks, directly fitting recorded trajectories is insufficient for motions involving rich contacts and lacks adaptability to different environments. Furthermore, strategies for fall damage reduction and recovery may differ between humans and robots due to differences in morphology and actuation. To address these issues, we formulate the prior learning problem as a {\textit{sparse key-frame tracking task in a goal-conditioned RL learning form}}, which can provide safety-critical posture anchors while leaving flexibility for the RL policy to explore and optimize its behaviors according to the robot’s own dynamics.  

The goal of this stage is thus to track a sequence of sparse key-frames, reach a final standing configuration, and minimize fall damage throughout the process. We formulate it as a finite-horizon control problem with horizon \(H=10\,\text{s}\), deliberately chosen to exceed the length of any collected demonstration, so that the robot must not only follow the motion but also maintain a stable standing posture afterwards. Formally, we define a sparse trajectory as  
$
\mathcal{P} = \{P_1, P_2, \dots, P_N, P_{\text{stand}}\},
$  
where each frame \(P_n\) is represented by  
$
P_n = \big\{ q_{i,n},\; \dot{q}_{i,n},\; T_{b_i,n},\; T_{\text{root},n},\; V_{b_i,n},\; V_{\text{root},n} \big\},
$  
with joint states \((q_{i,n}, \dot{q}_{i,n})\), root and link poses \((T_{\text{root},n}, T_{b_i,n})\), and corresponding twists \((V_{\text{root},n}, V_{b_i,n})\), sampled at a fixed frequency \(f\). The control objective requires the robot to reach each successive key-frame \(P_{n+1}\) within the interval \([t_n, t_{n+1}]\), and finally hold the standing frame \(P_{\text{stand}}\) until the episode terminates. For each trajectory, we train a corresponding augmentation policy. The training details are outlined as follows.

\noindent\textbf{State Initialization and Rollout.}
At the beginning of each episode, we randomly pick a frame \(P_0\) from the {\textit{dense}} trajectory, and copy all the joint and root information to {\textit{initialize}} the robot's state, while using only sparse key-frames for motion tracking. To simulate diverse fall conditions, we then randomize the base and joint states slightly and disable the actuators for a random duration \(t \sim \mathcal{U}(0.04, 1.0)\,\text{s}\), allowing the robot to enter free fall before regaining control.  

As the control and video frequency are not the same, we linearly interpolate all the information for the time between two frames. The final standing pose is set to be G1 robot's default pose with root height at 0.8m. The height is slightly higher than the robot’s actual standing height of \(\sim 0.728\,\text{m}\). This margin encourages the policy to stand more upright and avoid slouched postures during recovery. We keep the root yaw of the last frame the same as the trajectory's end, avoiding unnecessary rotation to a world-neutral orientation. Since retargeted trajectories may drift in root position, occasionally placing the robot above or below the ground, we preprocess each frame using forward kinematics and shift the root height by the lowest Z-coordinate among all rigid bodies, followed by a $0.05m$ offset to ensure clearance. This adjustment does not introduce harmful discontinuities, as the policy only follows sparse key-frames rather than dense trajectories, and in fact improves exploration between frames.

\noindent\textbf{Policy Optimization.} We optimize our policy using an asymmetric actor-critic framework with PPO~\cite{peng2018deepmimic,peng2022ase}. In this design, the critic has access to privileged information from the simulator that is unavailable to the actor and real-world setting. The actor’s observation space consists of: 1) root angular velocity $\omega_{root}$, 2) joint position and velocities $q, \dot{q}$, 3) last actions $a_{t-1}$, 4) joint position difference with respect to next key frame $q - q_{key}$ and 5) phase $\phi$. The critic network additionally observes root linear velocity $v_{root}$. The phase $\phi$ is calculated by dividing the current runtime $t$ by the total length of the trajectory in time $T$ and clipped to 1 for any time steps exceeding the trajectory length: $\phi = \text{min}(t/T, 1)$. 

\noindent\textbf{Domain Randomization.}
To further enhance the robustness of our policy and for later sim-to-real deployment, we followed previous works~\cite{huang2025learning} to adopt domain randomization during our policy training. We randomize the friction ($\mathcal{U}(0.25,1.75)$), payload ($\mathcal{U}(-1, 1)$), and gains for each joint (p-gain:$\mathcal{U}(0.9, 1.1)$, D-gain: $\mathcal{U}(0.9, 1.1)$), and randomly push the robots. We trained our robots on rough terrains to make our policy more robust in various environments. Observation noise is also added in simulation.

\noindent\textbf{Episode Termination.}  
As this task is contact-rich, we do not terminate episodes upon collisions, except when joint or root velocity limits are exceeded. In many locomotion tasks, episodes are typically terminated when the base height drops below a threshold or when collisions occur, to ensure the robot remains in valid poses. These criteria are unsuitable here, since our setting explicitly requires the robot to fall to the ground and recover. Likewise, motion-tracking tasks~\cite{cheng2024expressive} often use termination based on deviations from reference trajectories; however, because we only track very {\textit{sparse}} key-frames and aim to encourage exploration between them, this signal is also inappropriate. Therefore, our episode termination is kept minimal, while safety and stability are instead encouraged through the reward design described below.

\noindent\textbf{Rewards.} We formulate our rewards design mainly in 3 categories: 1) \textit{Tracking rewards}: These are main task rewards that track the difference between the current robot states and key-frame robot states, including joint positions and velocities, and rigid body poses and twists. Different from the trajectory information, where rigid body poses and twists are in local frame with respect to the robot root, here we calculate these quantities in world frame, which seamlessly integrates the tracking of root poses and twists as well. We calculate the reward using the function $h(d; \sigma) = \exp(-d^2 / \sigma)$. 2) \textit{Style rewards}: To penalize harmful and un-natural behaviors, we add this set of rewards to constrain on action rate, joint acceleration, torque values and out-of-limit joint behaviors. 3) \textit{Fall damage reduction rewards}: In order to mitigate the damage when falling on the ground, we add penalizing rewards on body collision, momentum change, and body yank as described in~\cite{ma2023learning}. The scale and exact definition of the rewards can be found in Table~\ref{tab:rewards}. In contrast to conventional fall recovery methods, our policy can utilize the safe motion pattern priors from human demonstrations to constrain fall recovery learning without complicated reward designs. 
\begin{table}[t]
\caption{Reward Terms Summary for Prior Learning. \textcolor{blue!60!black}{Tracking} /
\textcolor{orange!85!black}{Style} /
\textcolor{green!60!black}{Fall-damage reduction} rewards. }
\label{tab:rewards}
\renewcommand\arraystretch{1.2}
\resizebox{\columnwidth}{!}{
\begin{tabular}{l|l|r}
\toprule
\textbf{Reward Term} & \textbf{Definition} & \textbf{Scale} \\
\midrule 
\cellcolor{blue!10}Rigid body position tracking & \cellcolor{blue!10}$h\big(\sum_B w_B({T}_{B, w} - \hat{{T}}_{B, w})^2; \sigma\big)$ & \cellcolor{blue!10}1.25\\
\cellcolor{blue!10}Rigid body rotation tracking &\cellcolor{blue!10}$h\big(\sum_B ({R}_{B, w} - \hat{{R}}_{B, w})^2; \sigma\big)$ &\cellcolor{blue!10} 0.5\\
\cellcolor{blue!10}Rigid body linear velocity tracking & \cellcolor{blue!10}$h\big(\sum_B ({v}_{B, w} - \hat{{v}}_{B, w})^2; \sigma\big)$ &\cellcolor{blue!10} 0.125\\
\cellcolor{blue!10}Rigid body angular velocity tracking &\cellcolor{blue!10}$h\big(\sum_B ({\omega}_{B, w} - \hat{{\omega}}_{B, w})^2;\sigma\big)$ & \cellcolor{blue!10}0.125\\
\cellcolor{blue!10}Joint position tracking & \cellcolor{blue!10}$h\big(\sum_j (q_j - \hat{{q}}_{j})^2; \sigma\big)$ & \cellcolor{blue!10}0.5\\
\cellcolor{blue!10}Joint velocity tracking &\cellcolor{blue!10}$h\big(\sum_j (\dot{q}_j - \hat{\dot{q}}_{j})^2; \sigma\big)$  &\cellcolor{blue!10} 0.125\\

\cellcolor{orange!10}Joint position limit      &   \cellcolor{orange!10}$\sum_j \max(0, |q_j-q_j^{\text{limit}}|)$ &\cellcolor{orange!10} $-10$ \\
  \cellcolor{orange!10}Joint velocity limit      &   \cellcolor{orange!10}$\sum_j \max(0, |\dot{q}_j-\dot{q}_j^{\text{limit}}|)$ &  \cellcolor{orange!10}$-5$    \\
  \cellcolor{orange!10}Action rate &  \cellcolor{orange!10}$\sum (a[t] - a[t-1])^2$ &  \cellcolor{orange!10}  $-1e^{-3}$ \\
  \cellcolor{orange!10}Torques             &   \cellcolor{orange!10}$\sum_j \tau_j ^2 $ &  \cellcolor{orange!10}$-1e^{-6}$      \\
  \cellcolor{orange!10}Acceleration        &  \cellcolor{orange!10}$\sum_j \ddot{q}_j^2$ &   \cellcolor{orange!10}$-2.5e^{-7}$  \\
\cellcolor{green!10}Body collision        & \cellcolor{green!10}$\sum_{B} \lVert \lambda_B \rVert^2$   &\cellcolor{green!10} $-1e^{-7}$ \\
\cellcolor{green!10}Momentum change       & \cellcolor{green!10}$\sum_{B} \lVert m_B a_B\rVert$               & \cellcolor{green!10}$-5e^{-3}$ \\
\cellcolor{green!10}Body yank             & \cellcolor{green!10}$\sum_{B} \lVert \dot{F}_B\rVert^2 $        & \cellcolor{green!10}$-2e^{-6}$ \\
\bottomrule
\end{tabular}
}
\vspace{-4mm}
\end{table}
\subsubsection{Post trajectory stitching scheme} In real-world scenarios, losing balance does not always lead to a complete fall, as humans often adjust themselves and quickly regain stability. However, our human demonstrations only cover trajectories where a fall actually occurs. Training and expanding solely on such data would cause the robot to treat any minor imbalance as a full fall, leading to overly conservative while risky behavior. To address this, we propose to reuses demonstration trajectories with shortcuts, to generate alternative balance-preserving rollouts. The assumption behind this is that the robots can regain balance under a range of perturbations as long as a suitable reference key-frame can be found and to be used as the anchor, {\textit{i.e.},} it corresponds to a feasible intermediate state observed in recovery phases. Therefore, instead of explicitly training the policy on every such near-balance trajectory, we construct them via stitching and allow the robot to follow these recomposed trajectories at test time. Concretely, for a randomly selected $t < t_0$ (with $t_0$ set to around one third of the trajectory length), we create a shortcut to a later key-frame $t'$, selected from the second half of the trajectory, whose root height $h_{t'}$ is closest to the root height $h_t$ and satisfies a clearance threshold of $0.05$\,m. The policy is then re-executed from $s_t$ toward this new goal at $t'$, producing a stitched, new trajectory:
$G^{\text{new}} = \{(s_0, a_0), ..., (s_t, a_t), (s_{t'}, a_{t'}), ...,\}$. This procedure allows trajectories to be recomposed beyond their original temporal order, encouraging the policy to connect more arbitrary trajectory states with later recovery strategies. 

\subsection{Adaptive Memory Learning }\label{method-adapt}

\subsubsection{Distilling Priors via Diffusion Model} 

We use the expert policies and the post trajectory stitching scheme to collect $4.5$ million trajectory data pairs in the form of $(o, g, a)$, where $o$ denotes observations, $g$ represents reference sparse key-frames as goals, and $a$ are the corresponding actions. We expect to distill these diverse trajectories into a single policy. Since the distribution of these pairs is inherently multi-modal, directly fitting a unimodal policy would collapse diverse strategies into averaged behaviors, leading to unnatural motions and degraded safety. To preserve this multi-modality and further encourage variation, we adopt a diffusion policy\cite{huang2025diffuseloco} as a generative prior over trajectories. We keep a history of observations and goals to predict the next $H=12$ horizon of the actions, while only take the first action during inference time. By learning future steps of actions, the model can learn better transitions and relationship within histories of observations, and help predict the next-step action. Each observation and goals are embedded, with positional embedding and diffusion timestep embedding as well. A causal mask is used for attention computation, which means that action $a_t$ in the horizon can only have access to the information up to time at $t-1$. 

\subsubsection{Adaptive Goal Mapping}

\begin{figure}[t]
	\includegraphics[width=\columnwidth]{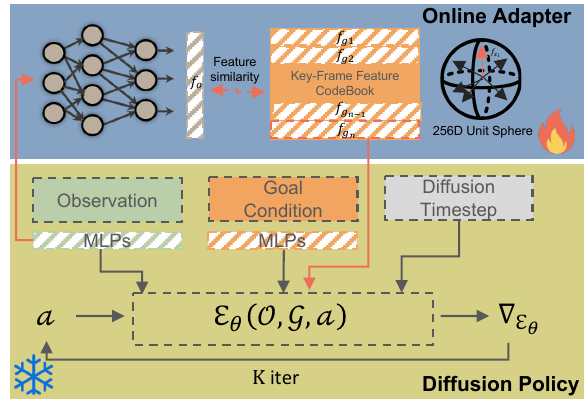}
	\vspace{-4mm}
	\caption{\textbf{Overview of online adapter.} During inference, the adapter uses the history of observations to dynamically predict a feature and match with a key-frame goal feature in the code-book, and then pass the matched goal feature into the diffusion model to guide the process with context-awareness.}
    \label{fig:adapt-overview}
	\vspace{-5mm}
\end{figure}

During training, key-frame goals are given and fixed according to the trajectories. However, in test time, the model cannot have access to which trajectory it needs to follow. Also, fixing goal sequences according to existing trajectories is not optimal especially in an environment that is different from training. To overcome this limitation, we introduce an online adapter as an MLP that dynamically adjusts key-frame conditions according to the current and past observations. This adapter will use a fixed-length history of observations to predict a feature vector which lies in the embedded space of the goal condition in diffusion models, as illustrated in Fig.~\ref{fig:adapt-overview}. We performed normalization on this embedded space to make it a unit sphere. A key-frame feature codebook 
$
\mathcal{F} = \{ f_{g_1}, f_{g_2}, \dots, f_{g_n} \}, with \quad \|f_{g_j}\|_2 = 1,
$
is pre-constructed from the augmented key-frames by passing into the fixed goal condition encoder in the diffusion model, where each entry stores the encoded feature $f_{g_j}$. 

During inference, for every 5 steps of action, the adapter will predict a feature given the observations, and retrieve the most relevant feature $f_g$ in the key-frame feature codebook by cosine similarity:  
$
j^* = \arg\max_{j} \; \frac{f_o^\top f_{g_j}}{\|f_o\|_2 \, \|f_{g_j}\|_2}, 
with \quad f_g = f_{g_{j^*}}.
$
By combining feature similarity with a normalized codebook on unit sphere, we ensure scale-invariant matching, preventing large feature norms from dominating and biasing reference selection. The selected feature $f_g$ is then provided to the diffusion policy, replacing the static trajectory input with a context-aware reference. This retrieval process ensures that policy conditions adapt in real time to robot’s state, while preserving safety through grounding in human priors.

\section{Experiments}

\subsection{Experimental Settings}
\noindent\textbf{Implementations.} We trained our first-phase sparse-keyframe policy in IsaacGym\cite{makoviychuk2021isaac}. The actor and critic network is a 2-layer MLP with hidden layer dimension [512, 256]. We trained in parallel with 4096 environments on Nvidia 4090 GPU for 5000 iterations per policy, which takes around 5 hours. We train our diffusion model for 1000 epochs, which takes around 40 hours on a single GPU of Nvidia A40. The robot we deploy on is 23-dof Unitree G1.

\noindent\textbf{Metrics.} We evaluate models based on three core criteria with levels of granularity: \locul{\textit{goal completeness}}, {\shareul{\textit{safeness}}}, and {\navul{\textit{efficiency}}}. {\textbf{1)}} For fall damage mitigation, we follow the evaluation criteria in~\cite{ma2023learning}, focusing on {\shareul{\textit{safeness}}} with the use of peak instantaneous impulse on the base (PII), mean base acceleration (BA), and peak joint internal forces across all joints (PIF). Since damage in heavy robots most often arises from high-impact stresses resulting from impulsive load transfer to the drives, higher values of these metrics indicate greater risk of damage and lower safety. {\textbf{2)}} For fall recovery, we consider all three dimensions -  a)~\locul{\textit{Goal Completeness}}: measured by the success rate (SR, $\%$), {\textit{i.e.,}} the percentage of episodes where the robot’s base height exceeds target height $0.7m$ and the robot remains upright for a sustained duration;
b) {\shareul{\textit{Safeness}}}: measured by time-to-fall (TTF, seconds), which evaluates stability based on how long the robot can remain standing before another fall occurs.
c) {\navul{\textit{Efficiency}}}: measured by the time-to-stand (TTS, seconds), evaluating the duration required for the robot to return to a stable height. Unless specified, simulated experiments are conducted with $512$ randomly spawned robots over 7.5s on uneven terrains, with randomized initial fall configuration, base mass, and noisy observations. Results are averaged over 5 runs to minimize random biases and verify robustness.

\subsection{Fall Damage Mitigation}\label{exp-fall}

\noindent \textbf{Settings.} As current research does not directly support fall damage mitigation on humanoid robot G1, we implement three baselines to compare with FIRM: {\textbf{1)}} {\textit{freezing model}}, where the robot output zero torque during falling, resulting in a passive collapse; {\textbf{2)}} {\textit{dense keyframe tracking model}}, where the robot follows dense keyframe references extracted from demonstrations; and {\textbf{3)}} {\textit{sparse keyframe tracking policy}}, where the robot follows sparse keyframe references extracted from demonstrations with only tracking rewards. This comparison setup allows us to investigate several key factors essential for fall damage mitigation: the passive vs active control strategies, the function of human demonstration, and the effect of adaptive memory for robust behaviors across diverse falling conditions.


\begin{figure}[t]
    \centering
    \subfigure[]{
        \includegraphics[width=0.45\linewidth]{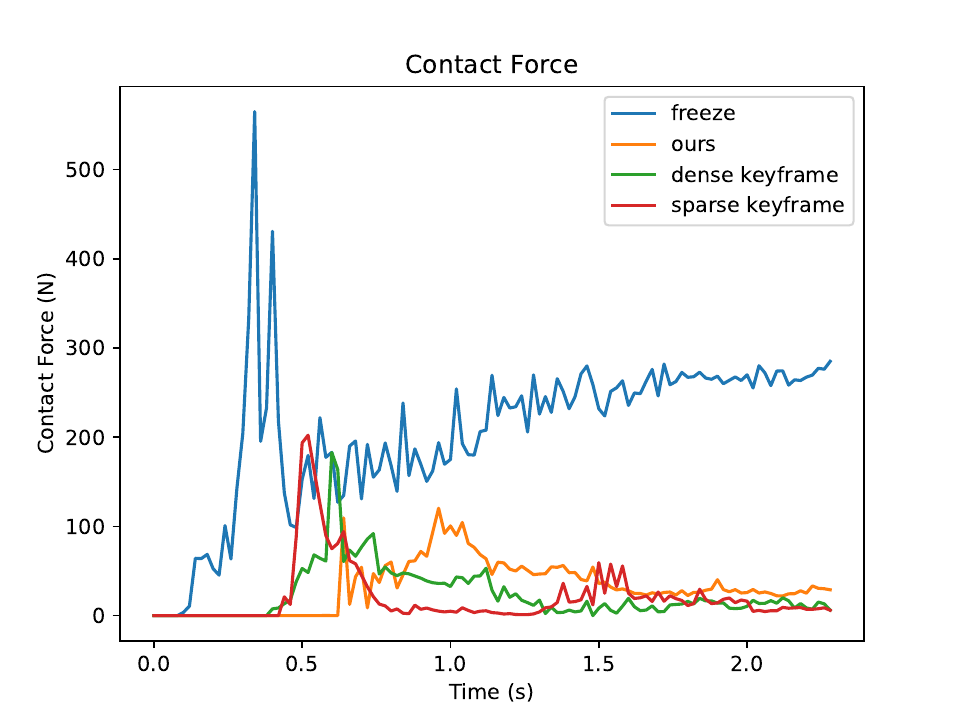}
        \label{fig:impulse}
    }
    \subfigure[]{
        \includegraphics[width=0.45\linewidth]{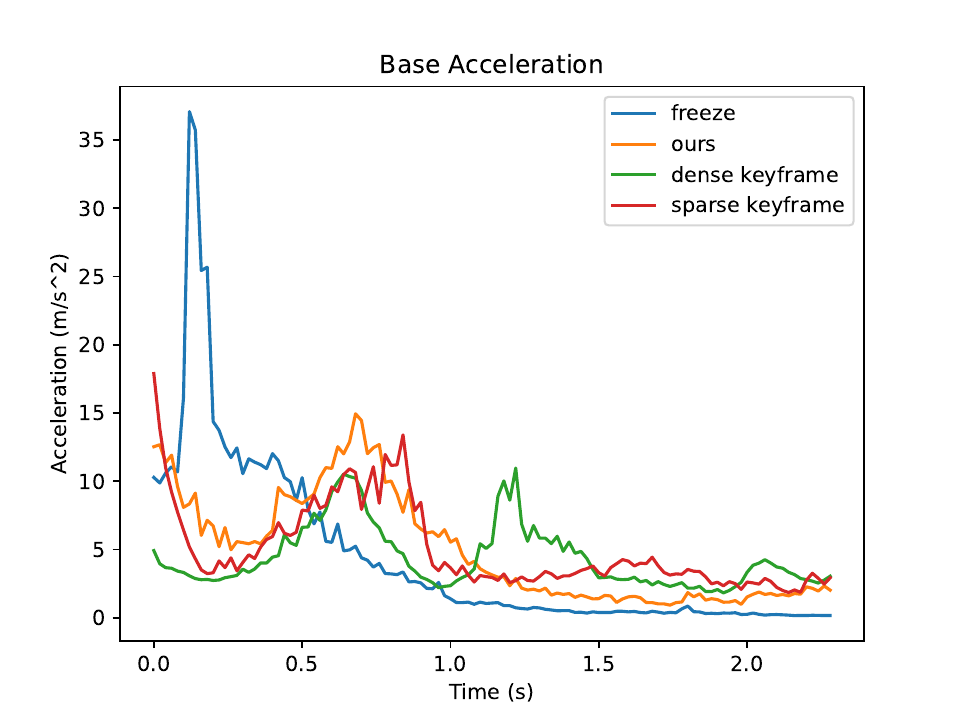}
        \label{fig:accel}
    }
    \caption{(a) Distribution of contact force on the base over all time steps. 
    Time steps with base contact impulse below $0.05\,\mathrm{Ns}$ are not included. 
    (b) Base acceleration (BA) during the fall.}
    \vspace{-2mm}
    \label{fig:fall_dynamics}
\end{figure}


\noindent \textbf{Results.}
We observed several key observations from Fig~\ref{fig:impulse} and Fig~\ref{fig:accel}: {\textbf{1)}} the freezing model yields the highest accelerations and joint forces, and its impulses are also the largest with multiple peaks (exceeding 400N), indicating that purely passive collapse exposes the robot to severe impact stresses; {\textbf{2)}} Sparse and dense keyframe tracking reduce impact forces compared to freezing model, with ours full FIRM model performing best in both base impulse and base acceleration. The underlying cause is that guiding the robot to follow human keyframes alone is brittle when the fall deviates from demonstrated trajectories. In contrast, the augmentation benefits from damage-reduction rewards, learning to emerge energy-dissipating poses, refine contact timing and force distribution beyond human priors, while online adaptation adaptively provides a safe, target goal that suits for current state to reference. Together, FIRM achieves smoother impact absorption and reduced joint stress.








\begingroup
\renewcommand\arraystretch{1.2}
\begin{table}[t]
\centering
\caption{\textbf{Humanoid G1 Robot fall recovery results in simulated environments.} 
Comparison between \textit{HoST}~\cite{huang2025learning} and \textbf{Ours} across three scenes. \textbf{N/A refers to no failure cases as standing first and falling later.}}
\vspace{-2mm}
\resizebox{\linewidth}{!}{%
\begin{tabular}{p{1.36cm}<{\centering} l p{1.36cm}<{\centering}p{1.36cm}<{\centering}p{1.36cm}<{\centering}p{1.36cm}<{\centering}p{1.36cm}<{\centering}p{1.36cm}<{\centering}}
\toprule
Terrain & Method & SR$\uparrow$ & TTF$\uparrow$ & TTS$\downarrow$ \\
\midrule
\multirow{2}{*}{Flat}
& HoST~\cite{huang2025learning} & \textbf{99.40}\ci{0.89} & 0.06\ci{0.12} & \textbf{1.75}\ci{0.03}\\
& \textbf{FIRM (Ours)} & 96.29\ci{5.27} & \textbf{N/A} & 2.47\ci{0.90} \\
\midrule
\multirow{2}{*}{Uneven}
& HoST~\cite{huang2025learning} & 23.20\ci{3.93} & 1.87\ci{1.28} & 3.10\ci{0.22}  \\
& \textbf{FIRM (Ours)} & \textbf{93.20}\ci{2.59} & \textbf{1.94}\ci{0.95} & \textbf{2.86}\ci{1.09} \\
\midrule
\multirow{2}{*}{Wave}
& HoST~\cite{huang2025learning} &10.20\ci{2.68}  & 1.62\ci{1.04} & \textbf{2.08}\ci{0.08} \\
& \textbf{FIRM (Ours)} & \textbf{55.86}\ci{2.49}  & \textbf{1.89}\ci{1.06} & 2.37\ci{1.12} \\
\bottomrule
\end{tabular}
}
\label{table:real_results}
\vspace{-4mm}
\end{table}
\endgroup

\begin{table}[t]
\centering
\caption{\textbf{Success rate under different payload masses.}}
\vspace{-2mm}
\renewcommand\arraystretch{1.2}
\resizebox{\linewidth}{!}{
\begin{tabular}{l p{1.36cm}<{\centering}p{1.36cm}<{\centering}p{1.36cm}<{\centering}p{1.36cm}<{\centering}p{1.36cm}<{\centering}p{1.36cm}<{\centering}p{1.36cm}<{\centering}}
\toprule
Method & 10kg & 12kg & 15kg & 20kg \\
\midrule
HoST~\cite{huang2025learning}   & \textbf{78.40}\ci{3.58} & 61.00\ci{6.04} & 35.60\ci{4.67} & 5.00\ci{1.58} \\
\textbf{FIRM(Ours)}  & 75.60\ci{4.61} & \textbf{72.61}\ci{4.83} & \textbf{60.20}\ci{5.81} & \textbf{21.20}\ci{7.19} \\
\bottomrule
\end{tabular}
}
\vspace{-4mm}
\label{table:payload_succ}
\end{table}

\subsection{Fall Recovery}
\noindent \textbf{Settings.} FIRM is our final policy ({\textit{i.e.,}} diffusion policy with keyframe codebook aware adapter). We compare FIRM with HoST~\cite{huang2025learning}, a recent SOTA method for humanoid standing-up control. HoST learns standing-up motions from scratch using RL with a multi-critic architecture and curriculum-based training, where a separate policy is trained for each terrain. We assess both methods from two perspectives: a) robustness to external disturbances introduced by additional payloads, and b) robustness to varying terrains. For fairness, we re-implement HoST~\cite{huang2025learning} using its official codebase and evaluate it under same simulation setup as FIRM.

\noindent \textbf{Results.} Tab.~\ref{table:real_results} and Tab.~\ref{table:payload_succ} indicate several insights. For flat terrain, our results are comparable to HoST across all three metrics, and we observed in experiments that once FIRM stands up on flat terrain, it does not fall again; therefore, the TTF is reported as N/A. For challenging terrains, our performance substantially surpasses HoST. Specifically, FIRM achieves significantly higher success rates (70\% improvement on uneven terrain and 45\% on wave terrain), while also maintaining comparable TTS by one second on average. Moreover, HoST often fails completely when additional payloads are introduced, whereas FIRM maintains stable recoveries with only minor degradation. These results highlight that the adaptive keyframe memory and online goal remapping in FIRM are critical for scaling recovery behaviors beyond nominal training conditions.

\begingroup
\setlength{\tabcolsep}{3pt}
\renewcommand\arraystretch{1.2}
\begin{table}[t]
\centering
\caption{\textbf{Ablation study of fall damage reduction \& recovery in simulation.} 
Peak Internal Force (PIF, N).}
\vspace{-1mm}
\resizebox{\linewidth}{!}{
\begin{tabular}{l cccc}
    \toprule
    \multirow{2}{*}{Method} & 
    \multicolumn{1}{c}{Damage Reduction} 
    & \multicolumn{3}{c}{Recovery (Overall)} \\
    \cmidrule(lr){2-2}
    \cmidrule(lr){3-5}
    & PIF$\downarrow$ %
    & SR$\uparrow$ & TTF$\uparrow$ & TTS$\downarrow$ \\
    \midrule
    Dense Keyframe & 42.38\ci{18.85} & 84.19\ci{3.96} & 1.51\ci{0.70} & 3.09\ci{0.12}\\
    Sparse Keyframe  & 41.87\ci{21.04} & 89.20\ci{1.92} & 2.49\ci{1.17} & 2.98\ci{0.08} \\
    + Augmentation Policy & \textbf{41.01}\ci{17.80} & 93.20\ci{2.59} & 1.94\ci{0.95} & 2.86\ci{1.09} \\
    Diffusion w/o Adaptor & 43.07\ci{18.24} & 92.32\ci{2.33} & 3.21\ci{1.02} & 2.99\ci{0.67} \\
    \textbf{FIRM(Ours)} & 41.23\ci{17.47} &  {\textbf{94.10}}\ci{2.17}& 2.73\ci{1.42} & \textbf{2.41}\ci{1.23} \\
    \bottomrule
\end{tabular}
}
\label{table:ablation_damage_recovery}
\vspace{-2mm}
\end{table}
\endgroup

\begin{figure}[t!]
    \centering
    \includegraphics[width=\linewidth]{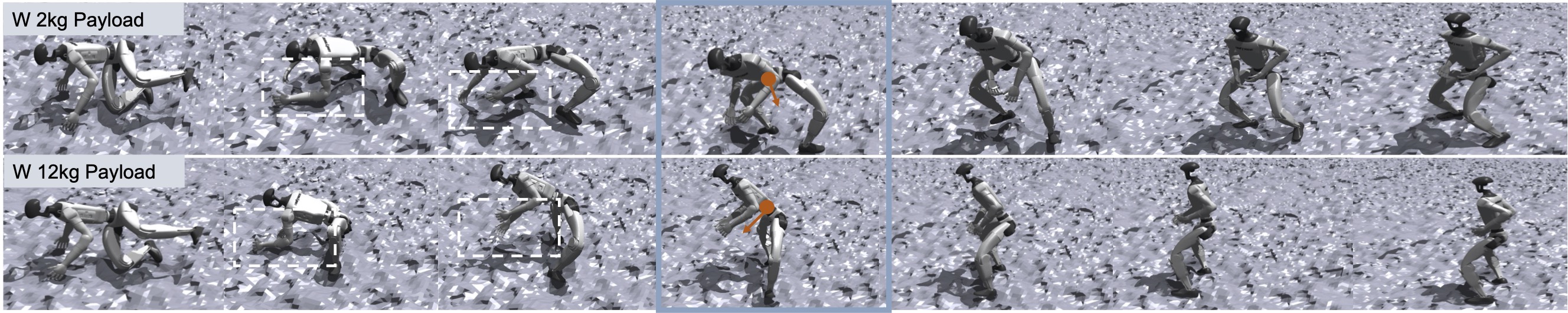}
    \vspace{-4mm}
    \caption{\textbf{Motion behaviors under different payloads.} As the white boxes show, For 2 kg payload, the arms perform a “support–push” motion. For 12 kg, the robot’s arms make full contact with the ground, exhibiting a forceful pushing action to lift the body. The orange arrow indicates torso orientation.}
    \label{fig:payload_sim}
\end{figure}

\subsection{Ablation Study}\label{sec:ablation}
\noindent \textbf{Settings.} 
To analyze the contribution of each component in FIRM, we conduct ablation studies on three simplified variants: {\textbf{1)}} \textit{Sparse Keyframe}, which directly tracks sparsely sampled human keyframes with only tracking rewards;  {\textbf{2)}} \textit{Sparse Keyframe + Augmentation Policy}, which augments demonstrations with RL and applies damage-reduction rewards; and 
{\textbf{3)}} \textit{Diffusion w/o Adaptor}, which distills multimodal fall-recovery strategies into a diffusion model but lacks the adaptive observation-to-goal adaptor at inference.  We compare these variants against our full method (\textbf{FIRM(Ours)}) under identical simulation settings, evaluating both fall damage reduction (PIF) and recovery performance (SR, TTF, TTS). Test terrains include \textit{flat}, \textit{uneven}, \textit{wave}, and \textit{rough}; among these, \textit{uneven} terrain is included during training, while the others are unseen test conditions.

\noindent \textbf{Results.} 
The ablation results are shown in Tab~\ref{table:ablation_damage_recovery}. It highlights the importance of each design choice in FIRM. Using only human demonstrations provides a better baseline: the robot learns safer fall behaviors compared with freezing, but recovery success remains limited and brittle under out-of-distribution states. Incorporating reinforcement-based augmentation rewards significantly reduces peak impact forces and base accelerations, showing that contact timing and redistribution can be optimized beyond human priors. However, without adaptive conditioning, recovery trajectories are still restricted by the original demonstration modes. Finally, equipping the policy with the adaptive keyframe memory and online adapter yields the most substantial gains. By dynamically remapping goals according to current states, the adapter enables the diffusion policy to generalize across terrains and payloads, achieving highest success rates with smoother recoveries.

FIRM demonstrates superior ability to minimize impact forces through context-aware adaptation and to generalize robustly across diverse fall directions.

\begin{figure}[t]
    \centering
    \includegraphics[width=\linewidth]{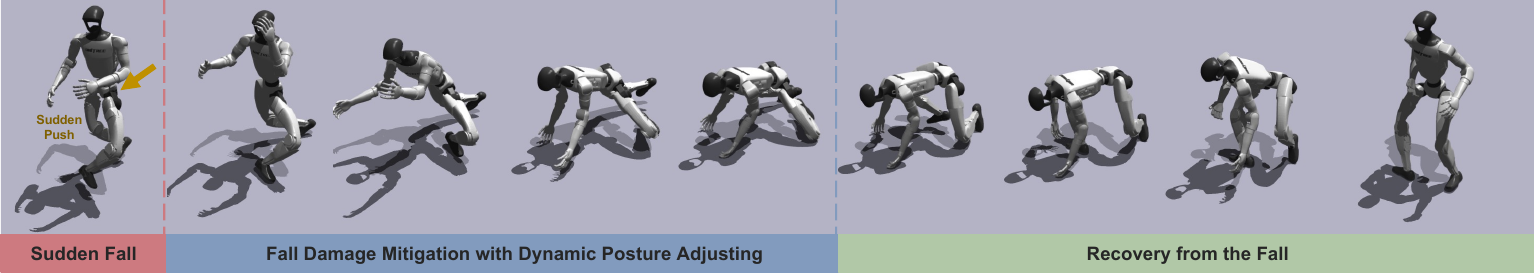}
    \vspace{-4mm}
    \caption{\textbf{Reaction to sudden fall.} After a sudden extra force, the robot try to spread arms to avoid base collision and push arms and knees to recover.}
    \label{fig:sudden_fall}
\end{figure}

\begin{figure*}[t!]
    \centering
    \includegraphics[width=\linewidth]{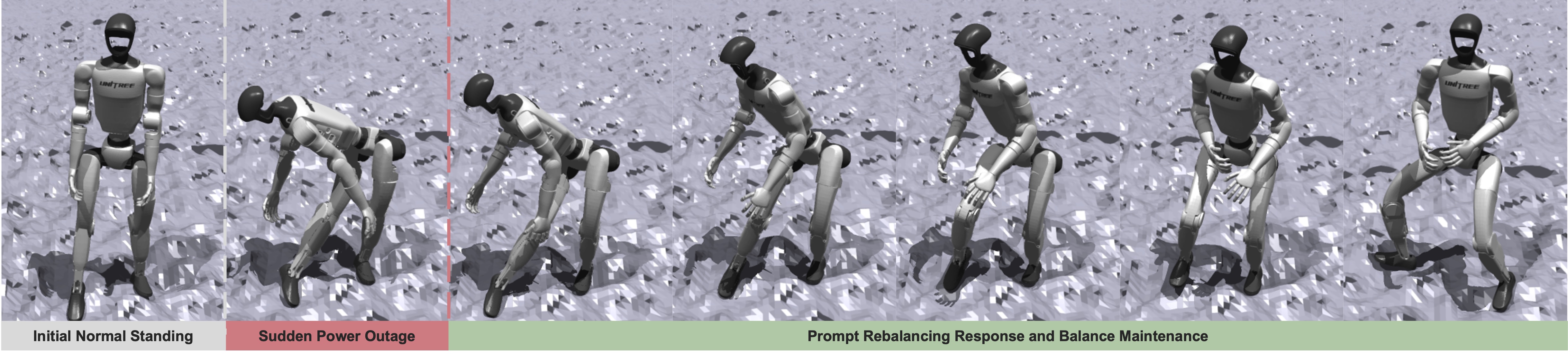}
    \vspace{-4mm}
    \caption{\textbf{Fall prevention.} Under 0.5s of zero torque output (mimics a sudden power outage in real world), the robot rapidly initiates rebalancing response, allowing the robot to maintain stability, thereby preventing a fall.}
    \label{fig:reblance-no-fall}
\end{figure*}

\begingroup
\begin{table}[h!]
\centering
\caption{Success rate (\%) for real-world fall recovery on the Unitree G1 across indoor terrains (10 trials each).}
\vspace{-2mm}
\resizebox{0.90\linewidth}{!}{%
\begin{tabular}{l c c c}
    \toprule
    Terrain & G1 Controller & HoST~\cite{huang2025learning} & \textbf{FIRM (Ours)} \\
    \midrule
    Flat Mat & \textbf{10/10} & 1/10 & \textbf{10/10} \\
    Slippery Surface & 7/10 & 0/10 & \textbf{8/10} \\
    \bottomrule
\end{tabular}
}
\vspace{-2mm}
\label{table:g1_success_real}
\end{table}
\endgroup

\begin{figure*}[t]
    \centering
    \includegraphics[width=\linewidth]{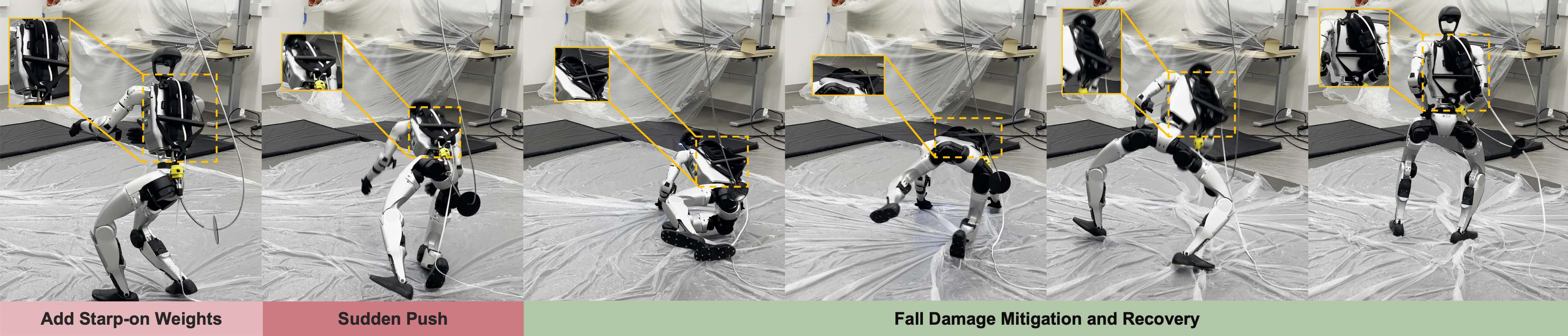}
    \vspace{-4mm}
    \caption{\textbf{Fall recovery with deformable payload on FIRM.} The robot achieves robust fall recovery while carrying a 2.7\,kg payload on both \textit{flat} and \textit{slippery} terrains.As highlighted in the yellow boxes, the payload visibly swings and deforms, introducing additional disturbances and further demonstrating the stability of our approach.}
    \label{fig:real-playload}
\end{figure*}

\begin{figure}[t]
    \centering
	\includegraphics[width=\columnwidth]{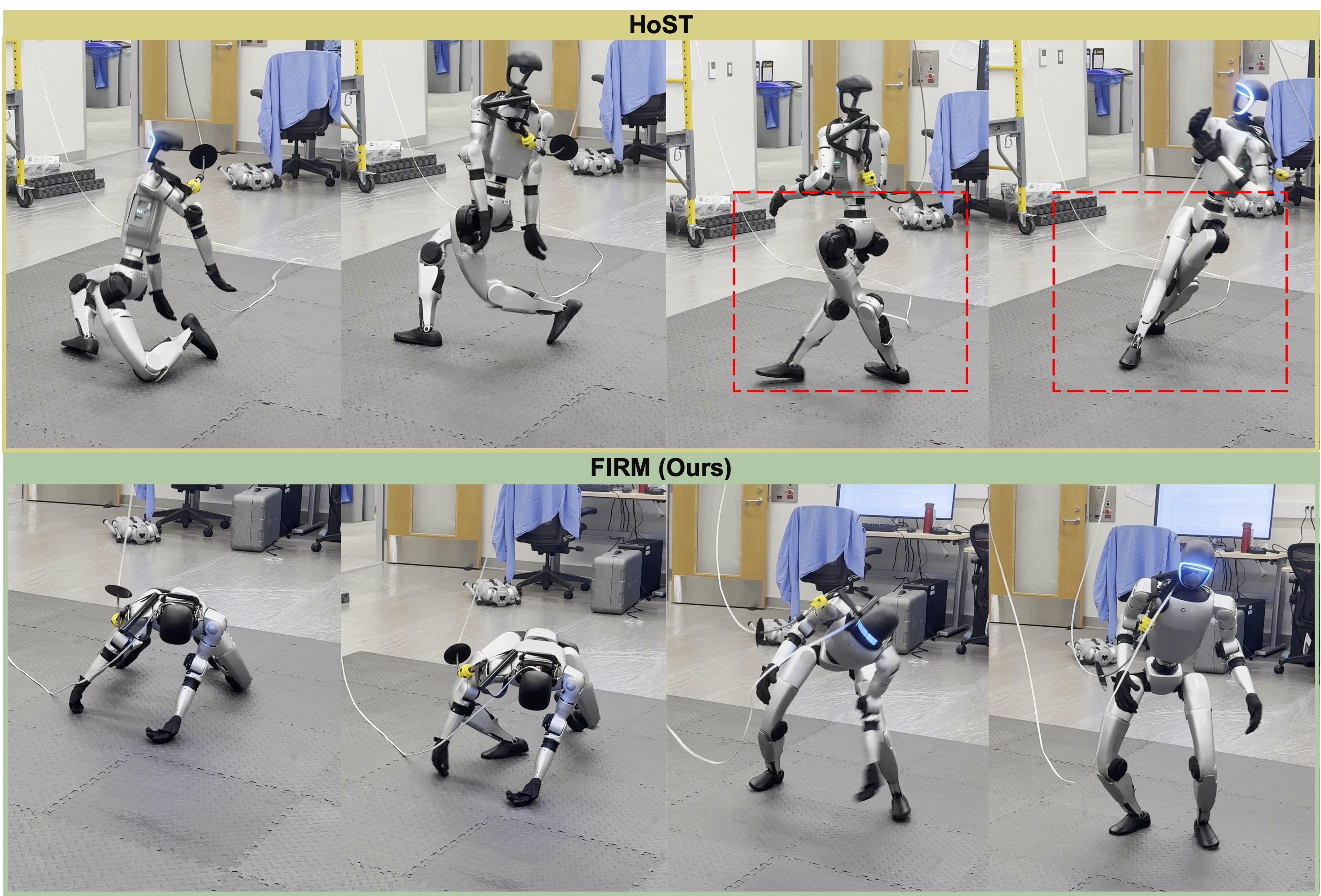}
    \vspace{-4mm}
	\caption{\textbf{Comparison of fall recovery in real-world deployment.} While HoST~\cite{huang2025learning} can barely stand, its leg movements are uncoordinated and the motion pattern departs from human-like behavior; Once upright, it fails to keep balance and quickly collapses. In comparison, our FIRM generates smoother and more natural motions, the robot uses arms to assist in a seamless fall-to-stand transition and sustain a stable posture after upright.}
    \label{fig:real_comp}
    \vspace{-4mm}
\end{figure}

\subsection{More Robustness Analysis}
We test FIRM under various settings to show its robustness. Fig.~\ref{fig:payload_sim} shows that under different payloads, FIRM adjusts its motion behaviors accordingly. Additional real world example can be seen at Fig.~\ref{fig:real-playload}. With a sudden external force making the robot fall inevitably, the robot can mitigate falling and recover as seen in Fig.~\ref{fig:sudden_fall}. Lastly, robot can regain balance successfully if a fall can be avoided (Fig.~\ref{fig:reblance-no-fall}).

\subsection{Real-world Comparison} 
\noindent \textbf{Settings.} We compare our FIRM against following baselines for fall recovery: {\textbf{1)}} \textit{G1 Controller}, the default manufacturer-provided recovery controller on the Unitree G1, which executes a hard-coded joint trajectory with PD stabilization; and 
{\textbf{2)}} \textit{HoST}~\cite{huang2025learning}, the recent state-of-the-art learning-based approach that trains standing-up policies from scratch using RL. We evaluate all methods on two in-door test terrains: a flat mat matrix and a slippery surface created by placing plastic sheets over the mat, as we observe that even these relatively simple conditions pose significant challenges to the baselines. The evaluation metric is the success rate measured over $10$ trials for each method and terrain.

\noindent \textbf{Results.} From Fig.~\ref{fig:real_comp} and Tab.~\ref{table:g1_success_real}, we observe that: 1) HoST fails to consistently stand up across all terrains. From the qualitative snapshots, we observe that the robot’s legs often cross during the recovery motion under HoST policy, leading to instability and preventing the robot from achieving a stable standing posture. 2) The G1 controller can stand up on both terrains, but since it is equipped with only a single predefined posture, it cannot generalize to diverse falling configurations.

\subsection{More Real-World Results}
We provide additional real-world experiments in the {\textit{supplementary video}} further to demonstrate the robustness of FIRM across diverse conditions, and investigate its limits. These include both indoor and outdoor terrains, variations in payload, and scenarios requiring active balance maintenance.

\section{Conclusions}

We present FIRM, the first learning framework that unifies fall mitigation and recovery within a single humanoid control policy.  
FIRM explicitly balances safety and behavioral diversity, embedding both throughout the learning process.
We highlight two key findings.
\textbf{1)} Safety can be grounded in human priors - a few key human poses provide seeds for safe falling and rising;  
\textbf{2)} Generalization and adaptiveness emerge from reinforcement-augmented priors, post-stitching of trajectories, and an adaptive key-frame codebook memory, enabling responsive recovery across diverse disturbances.
There remain two limitations.
{\bf 1)} FIRM depends on nearest-neighbour matching in its key-frame codebook, which may limit performance in highly out-of-distribution scenarios.  
{\bf 2)} It also relies solely on proprioceptive data and therefore cannot yet exploit external cues from vision or tactile sensing to enhance environmental awareness and contact safety.


\bibliographystyle{IEEEtran}
\bibliography{root.bib}
\clearpage
\appendix

\section{Appendix}
In this appendix, we provide: 1) additional implementation details (Sec.~\ref{sec:implementation}); 2) benchmark details (\Cref{sec:bench}); 3) further experimental results and analysis (\Cref{more_results}); and 4) a discussion on the broader impact/limitations/future work of our method (\Cref{more_discussion}).

\subsection{Implementation Details}\label{sec:implementation}

\noindent{\textbf{Retargeting Human Videos.}} For each recorded human motion video, we follow the pipeline of VideoMimic~\cite{videomimic} to retarget human motions onto the G1 humanoid. The videos are captured in 4K at 60 fps and downsampled to 30 fps for processing. We found that the default configuration of VideoMimic often fails to preserve the spatial relationships between body parts, leading to inconsistent limb coordination and distorted postures. In particular, the hip pitch and yaw joints frequently exhibit excessive rotations exceeding 180°–360° relative to their neutral poses in our scenarios. This limitation arises because VideoMimic is primarily designed for low-contact motions (\textit{e.g.}, walking, stepping, or sitting), whereas our sequences involve high-contact dynamics with frequent impacts during falls and recoveries. To mitigate this issue, we constrain the hip joint angle limits within physically plausible ranges that improve alignment between human and robot kinematics, resulting in more natural and structurally consistent retargeted motions.

There are also alternative frameworks to choose from, such as PHC~\cite{DBLP:conf/iccv/0002CWKX23}, and OmniRetarget~\cite{DBLP:journals/corr/abs-2509-26633} {\textit{etc.,}} imposing different physical constraints from different perspectives. However, our goal is not to achieve perfect physical fidelity, but to preserve the key motion intent in a physically reasonable manner, as we go beyond pure imitation that lacks contextual awareness, and instead use part of the retargeted motion sequence as a sparse prior. Residual inaccuracies and adaptation are then compensated through learning within the FIRM framework.

\noindent{\textbf{Training Details.}} All human demonstration videos are recorded on flat floors, while policy training for both stages (\textit{i.e.,} skill priors, and adaptive memory learning) is trained on {\textit{uneven}} terrains with domain randomization.

For trajectory collection used in distilling priors via the diffusion model, we record the following quantities: 1) root angular velocity $\omega_{root}$, 2) joint position and velocities $q, \dot{q}$, 3) last actions $a_{t-1}$. These quantities serve as the observation states for the diffusion policy, while the goal is represented by the joint positions of the corresponding target keyframe. For diffusion policy, we followed the transformer-based diffusion model as introduced in DiffuseLoco\cite{huang2025diffuseloco} with several architectural modifications: we employ a two-layer MLP to process the goal for enhancing its latent's representational capacity. The goal-state MLP consists of two layers of 128 neurons each and outputs a 64-dimensional latent embedding. To maintain representational consistency, we use the same MLP in a Siamese manner to encode the current joint positions, and compute the difference between the goal-encoded and current-encoded latents. This Siamese design mimics the joint-position-difference formulation used in our first-phase policy, allowing the diffusion model to reason about relative motion progress in latent space rather than relying solely on absolute joint configurations. 

The online adapter is implemented as a three-layer 1D convolutional network that processes the past 50 timesteps along the temporal dimension. It predicts a latent representation mapped to the same feature space as the goal's latent. The kernel sizes and strides for the three convolutional layers are [8, 4], [5, 1], and [5, 1], respectively. The output is normalized to lie on the unit sphere to ensure consistent feature magnitudes. The network is trained on the entire dataset for 20 epochs using a cosine similarity loss to optimize alignment between the predicted and target latents.
\subsection{Benchmark Details}\label{sec:bench}
\noindent{\textbf{Test Environments.}} While all human demonstrations are collected on flat floors, simulation training is performed on uneven terrains generated using Perlin noise. We test our approach in both simulation and real-world environments to evaluate robustness and generalization. 1) In simulation, we construct diverse terrains, including {\textit{flat, uneven, wave fields}}, and {\textit{slopes}}. We also design a special scenario with flat terrain and vertical walls to explicitly test fall-prevention behaviors when the robot falls toward obstacles. 2) In real-world experiments, we primarily test indoors with two levels of variation: friction and unevenness. For friction variation, we use different ground materials such as gym mats, plastic films, and protective wraps to create surfaces ranging from rough to slippery. For unevenness, we randomly scatter obstacles such as plastic foams and wooden planks to form irregular terrain patterns. Additionally, we evaluate outdoor performance on soft grass, earthen slopes, and sandy grounds resembling terrain on Mars, with varying inclines to assess stability and adaptability under natural conditions.

\noindent{\textbf{Baselines.}}
Since no existing methods directly address the same task as ours, most of our comparisons are conducted against ablated variants of our own model. For the fall-recovery task specifically, we additionally compare against the default G1 controller, and HoST~\cite{huang2025learning} that trains its policy purely through RL from scratch. We re-trained HoST following the official open-source implementation across multiple trials and observed notable discrepancies between our reproduced results and those reported in their paper. This issue was also discussed by other users as shown \href{https://github.com/InternRobotics/HoST/issues/14}{here}. As no further updates to the official code were available at the time we did the project, we used this retrained model for comparison. This outcome further highlights that learning fall recovery purely from scratch via reinforcement learning remains unstable and highly sensitive to reward design and environmental variations.
\subsection{More Results}\label{more_results}

\begingroup
\setlength{\tabcolsep}{3pt}
\renewcommand\arraystretch{1.2}
\begin{table}[h]
\centering
\caption{\textbf{Ablation on the number of keyframes.}
SR: Success Rate (\%), TTF: Time-to-Fall (s), TTS: Time-to-Steady (s), PIF: Peak Internal Force (N).}
\vspace{-2mm}
\resizebox{\linewidth}{!}{
\begin{tabular}{lcccc}
    \toprule
    \multirow{1}{*}{\# Keyframes} &
    \multicolumn{1}{c}{SR$\uparrow$} &
    \multicolumn{1}{c}{TTF$\uparrow$} &
    \multicolumn{1}{c}{TTS$\downarrow$} &
    \multicolumn{1}{c}{PIF$\downarrow$} \\
    \midrule
    Dense & 84.19\ci{3.96} & 1.51\ci{0.70} & 3.09\ci{0.12} & 42.38\ci{18.85} \\
    75 & 87.19\ci{2.57} & 0.62\ci{0.26} & 3.02\ci{0.14} & 41.98\ci{18.34} \\
    50 & 88.60\ci{3.29} & 2.92\ci{0.82} & \textbf{2.95}\ci{0.13} & 42.25\ci{20.92} \\
    25 & \textbf{89.20}\ci{1.92} & 2.49\ci{1.17} & 2.98\ci{0.08} & 41.87\ci{21.04} \\
    10 & 86.40\ci{3.36} & 2.16\ci{0.92} & 3.33\ci{0.18} & \textbf{41.39}\ci{21.02} \\
    \bottomrule
\end{tabular}
}
\label{table:ablation_keyframe}
\vspace{-2mm}
\end{table}
\endgroup
\begin{figure*}[t!]
    \centering
    \includegraphics[width=\linewidth]{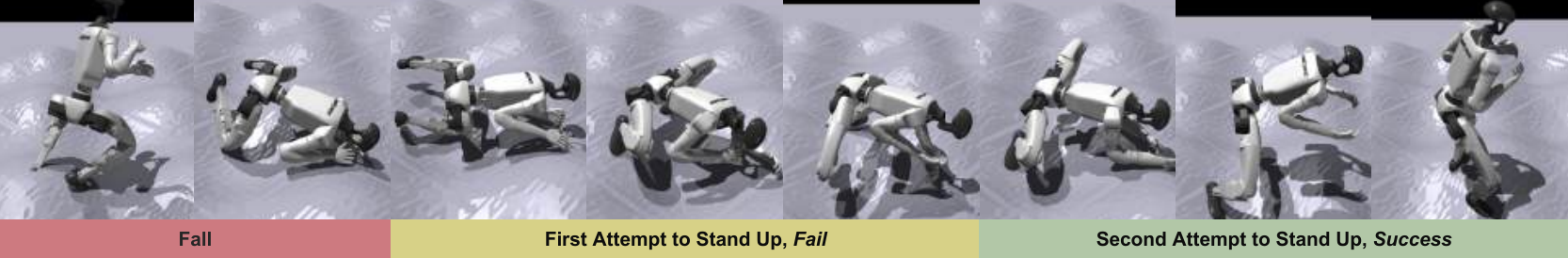}
    \vspace{-4mm}
    \caption{\textbf{Re-stand after a failed trial.} Instead of terminating upon a failed attempt, the online adapter dynamically re-evaluates the situation and identifies a new intermediate goal to initiate a subsequent re-stand trial.}
    \vspace{-2mm}
    \label{fig:restand}
\end{figure*}
\noindent{\textbf{Why Human Prior Need to Be Sparse?}}
To analyze the effect of keyframe density during Stage-1 skill prior learning, we conduct an ablation study using different numbers of keyframes per video. We also include a comparison with the \textit{Dense} keyframe setting, in which the policy follows interpolated motions between every frame, corresponding to an infinite number of keyframes in our human demonstrations. The test environments are kept identical to those described in Sec.~\ref{sec:ablation}. Each human demonstration lasts approximately five seconds, and we evaluate keyframe frequencies of 15 Hz, 10 Hz, 5 Hz, and 2 Hz, respectively.

The results are shown in Tab~\ref{table:ablation_keyframe}. We observe that both the success rate (SR) and time to stand (TTS) improve as the number of keyframes decreases, but drop sharply when the count falls below 25. This trend can be attributed to two main factors: 1) the retargeted human demonstrations and preprocessing are not perfectly accurate, making precise motion tracking across dense keyframes infeasible; and 2) dense keyframe tracking lacks adaptability to terrain variations that differ from the original capture conditions. Using fewer keyframes allows the policy greater freedom to adapt its motions to new terrains while maintaining overall trajectory consistency. However, when the keyframe count becomes too sparse, the robot struggles to infer appropriate intermediate motions, leading to degraded performance. Thus, the human prior should be sparse yet structured, providing sufficient temporal relationship guidance while allowing flexibility for environmental adaptation. Based on this trade-off, we adopt 25 keyframes as the default setting for our algorithm.

\noindent{\textbf{Supervised Finetuning, Na\"{i}ve Distillation (MLP), or Diffusion?}} There are multiple ways to integrate different skills or motion patterns into a unified policy. In this experiment, we evaluate two other major approaches. The first is \textit{Supervised Finetuning}, where we train a policy on one motion (motion 1), then sequentially fine-tune it on subsequent motions (motion 2, motion 3, etc.). The second is \textit{Distilling}, in which several expert policies are trained independently on different motions and subsequently distilled into a single policy using DAgger~\cite{ross2011reduction}. To ensure a fair comparison, the distilled policy network in \textit{Na\"{i}ve Distillation (MLP)} shares the same MLP architecture as all expert and fine-tuned policies. For \textit{Supervised Finetuning}, we train the initial motion for 5000 iterations and fine-tune on each subsequent motion for 3000 iterations. For \textit{Na\"{i}ve Distillation (MLP)}, we use DAgger to train the unified policy for 3000 iterations. Additionally, we include an \textit{Expert Policy} baseline that employs the Sparse Keyframe with Augmentation setup described in Sec.~\ref{sec:ablation}.

\begingroup
\setlength{\tabcolsep}{3pt}
\renewcommand\arraystretch{1.2}
\begin{table}[h]
\centering
\caption{\textbf{Ablation on the approach of integrating multiple motions.}
SR: Success Rate (\%), TTF: Time-to-Fall (s), TTS: Time-to-Steady (s), PIF: Peak Internal Force (N).}
\vspace{-2mm}
\resizebox{\linewidth}{!}{
\begin{tabular}{lcccc}
    \toprule
    \multirow{1}{*}{\# Keyframes} &
    \multicolumn{1}{c}{SR$\uparrow$} &
    \multicolumn{1}{c}{TTF$\uparrow$} &
    \multicolumn{1}{c}{TTS$\downarrow$} &
    \multicolumn{1}{c}{PIF$\downarrow$} \\
    \midrule
    Expert Policy & \textbf{93.20}\ci{2.59} & 1.94\ci{0.95} & \textbf{2.86}\ci{1.09} & \textbf{41.01}\ci{17.80} \\
    Supervised Finetuning & 30.50\ci{15.20} & 0.93\ci{0.20} & 3.90\ci{1.12} & 43.28\ci{17.32} \\
    Distilling MLP & 76.18\ci{5.23} & 1.45\ci{0.87} & 3.41\ci{0.94} & 43.36\ci{18.21} \\
    Diffusion w/o Adaptor & 92.32\ci{2.33} & 3.21\ci{1.02} & 2.99\ci{0.67} & 43.87\ci{18.24} \\
    \bottomrule
\end{tabular}
}
\label{table:ablation_motion_integration}
\vspace{-2mm}
\end{table}
\endgroup

The results can be seen from Tab~\ref{table:ablation_motion_integration}. We observe that the performance of \textit{Supervised Finetuning} degrades significantly compared to \textit{Policy Distillation (MLP)}. Sequential finetuning leads to catastrophic forgetting, causing the policy to retain only the most recently trained motion while losing earlier ones. While for \textit{Distilling} with MLP, the performance also drops. We see that the model works well on relatively similar motions, while performing worse on others, causing the overall results to decrease. As the lengths for different motions are different, it is challenging for the phase term to capture the correct next-step dynamics for each, and the simple MLP architecture cannot effectively model the multimodal nature of these motions. These observations highlight the necessity of a diffusion-based approach, which can distill motion priors across heterogeneous demonstrations while preserving multimodality.
\begin{figure*}[h!]
    \centering
    \includegraphics[width=\linewidth]{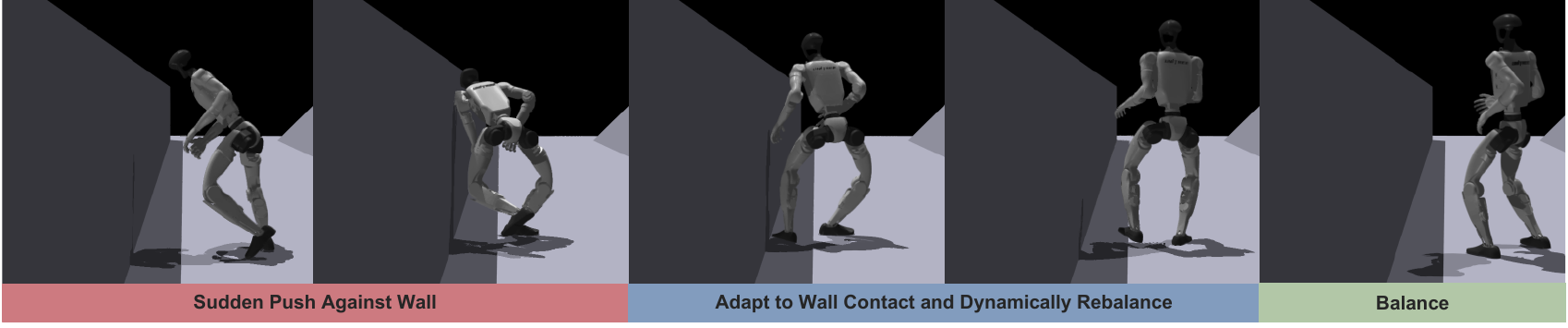}
    \vspace{-4mm}
    \caption{\textbf{Fall prevention against wall.} Upon a sudden push toward the wall, the robot adapts to the contact, using it as support to stabilize its posture and avoid falling.}
    \vspace{-2mm}
    \label{fig:wall}
\end{figure*}

\noindent{\textbf{The Importance of Online Adapter.}}
Online adaptation is crucial for a diffusion model to adapt in a dynamic environment where predefined motion sequences are infeasible. Rather than relying on a fixed trajectory, the online adapter serves as a motion planner that allows the robot to adjust its behavior according to the current state and surroundings. One significant outcome of our online adapter is the ability to re-stand from a previously ``failed" sequence, demonstrating improved contextual awareness and adaptability. An example of such re-standing behavior is illustrated in Fig~\ref{fig:restand}.

\begin{figure*}[h]
    \centering
    \includegraphics[width=\linewidth]{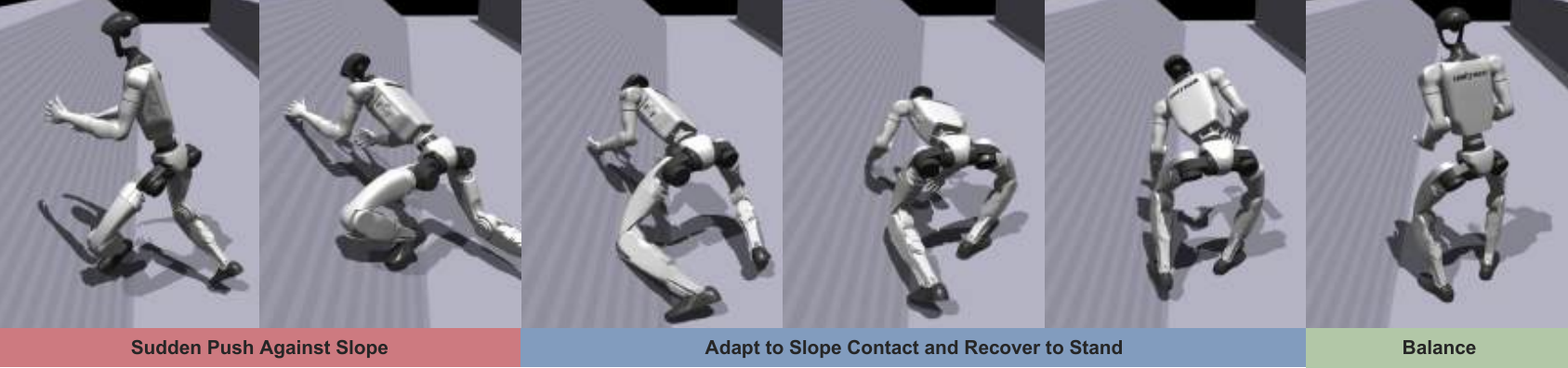}
    \vspace{-4mm}
    \caption{\textbf{Fall prevention against slope.} Similar to the wall scenario, the robot adapts to the slope contact to prevent a full fall, and leverages the inclined surface as external support to recover and subsequently stand upright with regained balance.}
    \vspace{-2mm}
    \label{fig:slope}
\end{figure*}

\noindent{\textbf{Additional Fall Prevention Results.}}
Additionally, we test the fall prevention performance against walls and slopes. Our robot demonstrates the ability to adapt to previously unseen environmental contacts, effectively leveraging surrounding structures to support itself and regain balance. Results in the simulation environment can be seen in Fig~\ref{fig:wall} and~\ref{fig:slope}.
\subsection{Discussion}\label{more_discussion}
FIRM goes beyond imitation learning or reinforcement learning alone, providing a viable pipeline for leveraging a small number of human demonstrations to solve complex, contact-rich tasks that are otherwise difficult to model through imitation or reward design to adapt in the dynamic environments.
By learning from sparse human demonstrations, our framework can generate context-aware responses in time through the online adapter, adjusting its next goal to achieve stable task execution. Still, several limitations and open challenges remain: 1) As the number of our demonstrations is small and they are very iconic and different from each other, we rarely observe cases where the robot combines the falling phase of one motion with the early stage of recovery strategy of another sequence, more near the end when adjusting its final standing pose. This suggests that while our model achieves effective interpolation within each sparse motion segment in the learned goal feature space, the latent representations of different motion sequences remain relatively disjoint, limiting cross-motion composition.  A promising direction for future work is to dig into more primitive and compositional motion structures, so that complex behaviors can emerge from more flexible recombination of simpler motor elements. Such a representation would enable smoother transitions and motion blending between goal features, allowing the robot to dynamically compose strategies across diverse situations under just a few demonstrations. 2) Human priors should extend beyond motion trajectories to include decision-making processes, \textit{i.e.}, understanding why humans choose specific recovery strategies under certain environmental conditions. Since our current demonstrations are recorded only on flat terrains, they involve limited decision complexity. Capturing first-person visual perspectives from humans performing falls across varied terrains could provide richer contextual cues, allowing robots to learn both perceptual grounding and adaptive decision-making in more dynamic environments. Integrating such perceptual and cognitive priors represents an exciting direction for future research.
\end{document}